\documentclass{article}
\usepackage{arxiv}

\usepackage[utf8]{inputenc} 
\usepackage[T1]{fontenc}    
\usepackage{url}            
\usepackage{booktabs}       
\usepackage{amsfonts}       
\usepackage{nicefrac}       
\usepackage{microtype}      
\usepackage{xcolor}         
\usepackage{wrapfig}
\usepackage{graphicx} 
\usepackage{enumitem}
\usepackage{amssymb}
\usepackage{amsmath}
\usepackage{amsthm}
\usepackage{pifont}
\usepackage{algorithm}
\usepackage{algorithmic}
\usepackage[normalem]{ulem}
\useunder{\uline}{\ul}{}
\usepackage[colorlinks,linkcolor=red,anchorcolor=blue,citecolor=green]{hyperref}
\usepackage{tcolorbox}
\usepackage{multirow}
\usepackage[authoryear]{natbib}

\title{\textsc{BrokenBind}: Universal Modality Exploration beyond Dataset Boundaries}

\author{
  Zhuo Huang$^{1}$,  
  Runnan Chen$^{1}$,   
  Bo Han$^{2,3}$,
  Gang Niu$^3$,
  Masashi Sugiyama$^{3,4}$,
  Tongliang Liu$^{1}$\\[1ex]
  \small{$^1$Sydney AI Centre, The University of Sydney;}
  \small{$^2$Hong Kong Baptist University;} \\
  \small{$^3$RIKEN Center for Advanced Intelligence Project;}
  \small{$^4$The University of Tokyo}\\
}

\date{}

\renewcommand{\shorttitle}{\textit{BrokenBind}}

\begin{document}
\maketitle

\begin{abstract}
	Multi-modal learning combines various modalities to provide a comprehensive understanding of real-world problems. A common strategy is to directly bind different modalities together in a specific joint embedding space. However, the capability of existing methods is restricted within the modalities presented in the given dataset, thus they are biased when generalizing to unpresented modalities in downstream tasks. As a result, due to such inflexibility, the viability of previous methods is seriously hindered by the cost of acquiring multi-modal datasets. In this paper, we introduce \textbf{BrokenBind}, which focuses on binding modalities that are presented from different datasets. To achieve this, BrokenBind simultaneously leverages multiple datasets containing the modalities of interest and one shared modality. Though the two datasets do not correspond to each other due to distribution mismatch, we can capture their relationship to generate pseudo embeddings to fill in the missing modalities of interest, enabling flexible and generalized multi-modal learning. Under our framework, any two modalities can be bound together, free from the dataset limitation, to achieve universal modality exploration. Further, to reveal the capability of our method, we study intensified scenarios where more than two datasets are needed for modality binding and show the effectiveness of BrokenBind in low-data regimes. Through extensive evaluation, we carefully justify the superiority of BrokenBind compared to well-known multi-modal baseline methods.
\end{abstract}


\vspace{5mm}
\section{Introduction}
\label{sec:introduction}
Multi-Modal Learning~\cite{ma2021smil, ngiam2011multimodal, xu2023multimodal, zhang2023meta,huang2025mllm,huang2025surprise3d,huang2025towards,liu2025towards} leverages different modalities to understand real-world scenarios from multiple perspectives, which provably outperforms uni-modal learning~\cite{huang2021makes}, meanwhile showing remarkable generalization performance~\cite{radford2021learning}. To combine diverse modalities and enrich information understanding, one common strategy is to \emph{bind} different modalities into one joint space~\cite{girdhar2023imagebind, wang2024freebind, zhu2023languagebind}. As a result, generalization across multiple modalities can be successful, thus benefiting various downstream tasks, such as information captioning~\cite{nguyen2024improving, zhang2022magic}, knowledge retrieval~\cite{jain2021mural, lin2020learning, zeng2023multi}, and cross-modal generation~\cite{liao2024text, zhang2021cross}.

\begin{wrapfigure}{r}{0.5\textwidth}
\centering
\vspace{-5mm}
\includegraphics[width=0.5\textwidth]{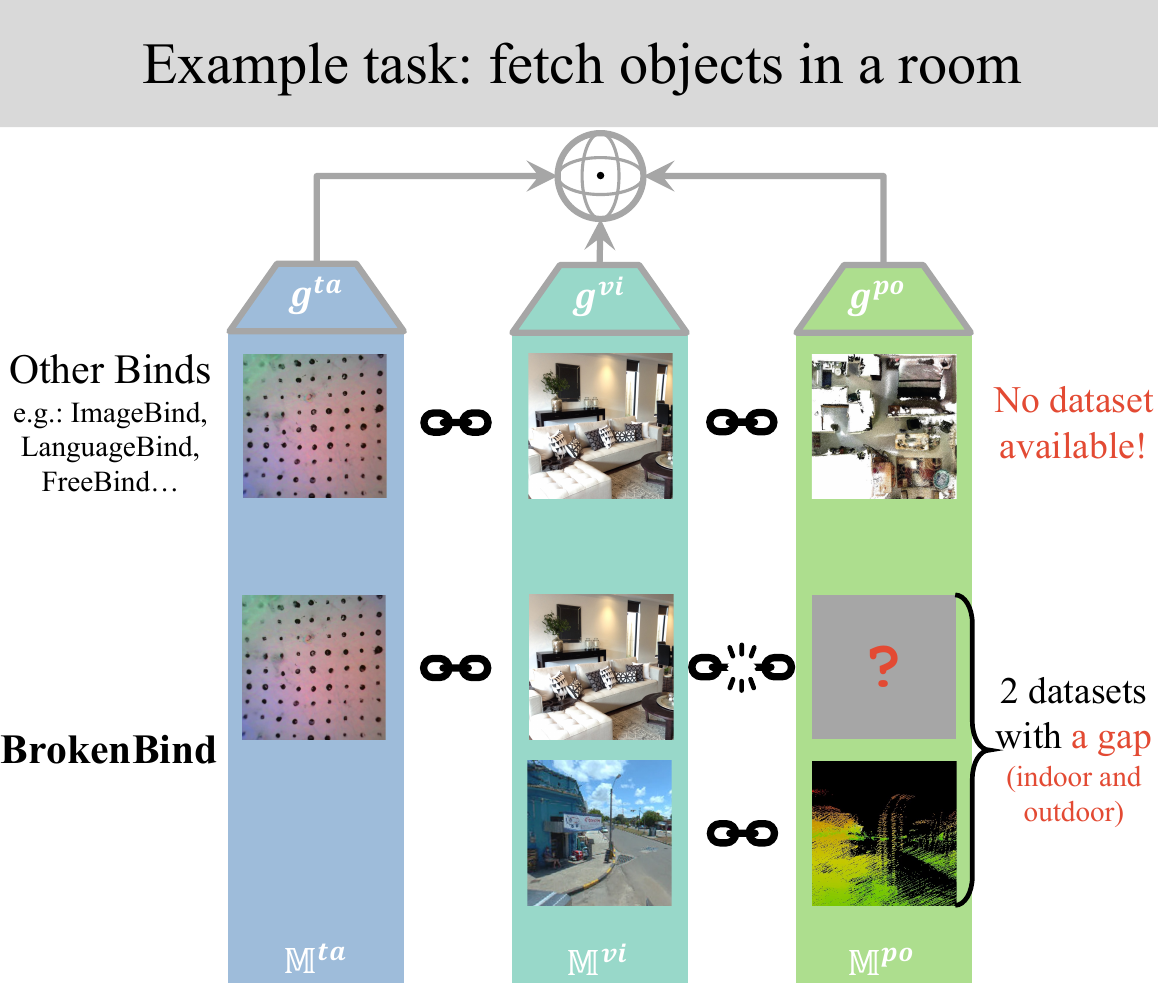}
\caption{Example of BrokenBind and previous binding strategy for fetching objects in a room. The three modalities tactile sensing (``ta''), vision (``vi''), and point cloud (``po'') are from space $\mathbb{M}^{ta}$, $\mathbb{M}^{vi}$, and $\mathbb{M}^{po}$, respectively, and are further correspondingly encoded by encoders $g^{ta}$, $g^{vi}$, and $g^{po}$ into a joint space (denoted by different colors). Existing approaches require multi-modal correspondence which is hard to satisfy, instead, BrokenBind aims to leverage multiple datasets effectively by bridging the dataset gap.}
\label{fig:problem}
\end{wrapfigure}

Existing modality-binding studies, such as ImageBind~\cite{girdhar2023imagebind}, LanguageBind~\cite{zhu2023languagebind}, and FreeBind~\cite{wang2024freebind} leverage the contrastive language-image pre-training (CLIP) framework~\cite{radford2021learning} to bind two modalities at a time, further using shared modalities to bridge all modalities in the same joint space. However, such a modality-binding strategy requires that: 1) the given dataset covers all modalities of interest, and 2) containing instance-level modality correspondence to pair different modalities together. In practice, collecting multiple modalities with correspondence is quite challenging and cost-expensive, which largely limits the flexibility of modality binding and hinders its feasibility in real-world applications. As shown in Figure \ref{fig:problem}: Suppose we instruct a multi-modal AI system to fetch objects in a room, for which three modalities are needed, namely tactile sensing of pillows (``ta''), visual understanding (``vi''), and point cloud of the room (``po''). The existing strategy would require collecting all three modalities and then conduct modality binding. Unfortunately, there is no existing dataset that fulfills all three modalities with instance-correspondence, which makes the binding result unreliable. In future AI, large-scale multi-modal training on multiple datasets would inevitably face such an issue where we only have limited desired modalities with serious distribution shifts~\cite{yamane2021mediated, yamane2023mediated}, \textit{e.g.}, one dataset is from indoor environments and another one is from outdoors. Consequently, they cannot be directly bound together, restricting the diversity of the joint embedding space and hindering the generalization across modalities, especially emergent alignment~\cite{girdhar2023imagebind}.

Targeting flexible and generalized modality binding, we propose \emph{BrokenBind}, an approach designed to accurately bind modalities that are not presented in the given dataset at the same time. Specifically, we can simultaneously leverage multiple datasets that contain the modalities of interest (\textit{i.e.}, target modalities), along with at least one shared modality as a pivot (\textit{i.e.}, pivot modality). In this way, it is possible to generalize through the pivot modality and achieve modality binding. This is similar to machine translation when translating from a minor language (\textit{e.g.}, Norwegian) to another minor one (\textit{e.g.}, Tibetan) by utilizing pivot languages (\textit{e.g.}, English and Chinese)~\cite{wu2007pivot, utiyama2007comparison}. During binding, we conduct \emph{modality extrapolation} to tackle the ``broken'' correspondence between two datasets as they are sampled from different distributions. Such an extrapolation process can be achieved based on two types of relationships: 1) cross-modal relationships from the pivot modality to target modalities, and 2) cross-data relationships of the pivot modality data from different datasets. By capturing such two relationships realized by transition matrices, we enable modality extrapolation to produce pseudo embeddings that can compensate for the target modalities missing from each dataset. Further leveraging the pseudo embeddings, we can create additional pairs to train our binding models.

Thanks to the modality extrapolation, the proposed BrokenBind framework inherently calibrates the distribution gap between different datasets. In this way, we can successfully enable generalization to modalities that are not presented in the current dataset, such as inferring the audio knowledge of a ``cat'' from vision dataset ImageNet~\cite{deng2009imagenet} and retrieving tactile sensing from a ``chair'' from 3D point cloud dataset ShapeNet~\cite{chang2015shapenet}. Moreover, any type of modality from the given target dataset can be explored, as long as we can collect proper support dataset. Therefore, the proposed BrokenBind achieves \textit{universal} modality exploration by leveraging knowledge of any possible dataset. The above achievement of the proposed BrokenBind is verified by extensive experiments on various datasets.

To sum up, our contributions are three-fold:
\begin{itemize}
    \item We reveal a realistic modality binding problem where modalities are not fulfilled by the given dataset and identify the limitations of existing methodologies.
    \item We propose BrokenBind to tackle such a problem by conducting modality extrapolation through capturing relationships across modalities and datasets.
    \item Extensive experimental justifications demonstrate the effectiveness of BrokenBind, providing an effective solution to real-world applications.
\end{itemize}

\section{Related Work}
\label{sec:related_work}

\paragraph{Modality Binding} aims to bind different modalities such that the knowledge of any modality can be retrieved by providing just one modality. Existing strategies are based on contrastive language-image pre-training (CLIP)~\citep{radford2021learning} framework which aligns image and text data through contrastive learning~\cite{he2020momentum}. ImageBind~\cite{girdhar2023imagebind} is the first work proposed to bind diverse modalities together by leveraging various datasets that all share vision modality. Therefore, all other modalities are aligned with the vision data. Interestingly, an intriguing effect named emergent alignment reveals the alignment of two modalities can be enabled by binding them to a third modality. Thanks to this, ImageBind successfully constructs a joint embedding space using the vision modality as a pivot. Inspire by ImageBind, LanguageBind~\cite{zhu2023languagebind} proposed a similar framework but using language as the center modality. Further, a more versatile bind named FreeBind~\cite{wang2024freebind} considers the knowledge extension of ImageBind and LanguageBind by incorporating additional knowledge space such as language and audio space CLAP~\cite{elizalde2023clap} or point-cloud and language space ULIP~\cite{xue2023ulip}. Recently, OmniBind~\cite{wang2024omnibind} and UniBind~\cite{lyu2024unibind} leverage large language models (LLMs)~\cite{touvron2023llama,huang2024machine,wang2024noisegpt} to extend the binding space.

\paragraph{Cross-Modal Generalization} aims to leverage the knowledge from given modalities and generalize to unknown modalities, which is similar to out-of-distribution (OOD) generalization~\cite{arjovsky2019invariant, muandet2013domain,hong2024improving,hong2024your,hong2025data,huang2024winning,huang2023harnessing,huang2023robust} but still differs from it due to the significant modality gap. Yet still, existing methodologies normally leverage the techniques from OOD generalization. Shen et al.~\cite{shen2023cross} proposed to conduct distribution alignment through a gradual process to progressively tackle the modality gap. Further, Cai et al.~\cite{cai2024enhancing} selectively replaces some active features from one modality to another one to conduct modality generation. By training with such generated features, cross-modal generalization can be achieved. Liang et al.~\cite{liang2021cross} considered a dynamic learning problem where novel modality data emerges for training, which is tackled by meta-learning. Xia et al.~\cite{xia2024achieving} proposed a unified representation method that encodes different modalities into the same space and trains them through semi-supervised learning \cite{huang2022they,huang2023flatmatch,li2024dynamic} with mutual information regularization. Recently, Liang et al.~\cite{liang2023multimodal} proposed a theoretical framework to discover the interaction between different modalities. By doing so, the generalization across modalities can be achieved by exploring the shared information.

\paragraph{Discussion:} Complementary to the above two branches of multi-modal learning, this paper takes advantage of existing binding strategies and explores the knowledge of unknown modalities. Specifically, due to the modality diversity and correspondence from the dataset, the emergence alignment of modality binding is still suboptimal, \textit{e.g.}, text-audio retrieval and vision-audio retrieval (mAP) results from ImageBind and LanguageBind is only $5.3\%$ and $6.2\%$ on AVE dataset~\cite{tian2018audio}, respectively. However, BrokenBind identifies the bias caused by mismatched datasets and compensates for the missing knowledge through modality extrapolation. As a result, modality binding can be calibrated by achieving $25.2\%$ and $19.6\%$ mAP under the same setting, thus it is capable of generalizing to unseen modalities.

\begin{wrapfigure}{r}{0.5\textwidth}
    \centering
    \vspace{-5mm}
    \includegraphics[width=0.5\textwidth]{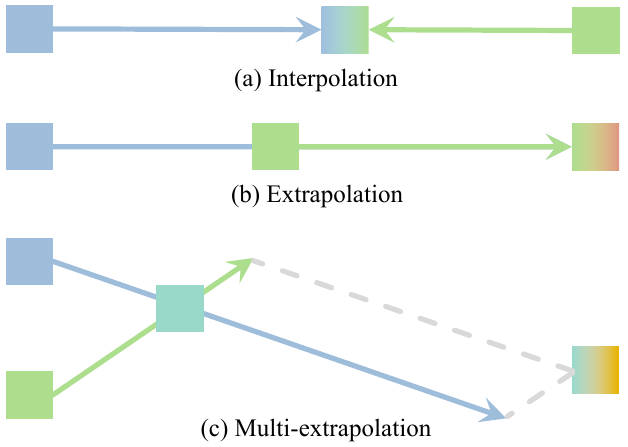}
    \caption{Illustration of interpolation, extrapolation, and multi-extrapolation. Different colors denote different examples. The interpolation shows the knowledge in two known examples, but the extrapolation show emerging knowledge.}
    \vspace{-8mm}
    \label{fig:extrapolation}
\end{wrapfigure}
\section{Methodology}
\label{sec:methodology}

In this section, we first formalize the setting of our BrokenBind. Then, we start from interpolation via a popular method Mixup~\cite{zhang2017mixup,huang2021universal} and extend it to extrapolation. Further, we demonstrate the methodology of BrokenBind in detail and explain the learning strategy based on a common assumption of cross-modal cross-dataset consistency.

\subsection{Problem Setting}
Under BrokenBind setting, suppose we are given two datasets $\mathcal{D}^1=\{(x^{a1}_i, x^{b1}_i)_{i=0}^{m}\}$ and $\mathcal{D}^2=\{(x^{b2}_j, x^{c2}_j)_{j=0}^{n}\}$, each of them contains two modalities as specified by the superscripts $a$, $b$ and $c$\footnote{Note that it can be extended to more than $2$ modalities and $2$ datasets.}. We denote that an example $x^{a}$ of modality $a$ is from space $\mathbb{M}^a$, and same for $\mathbb{M}^b$ and $\mathbb{M}^c$. The modality shared across different datasets is denoted as ``pivot'' modality, and the modality missing from each dataset is termed ``target'' modality. For example, the pivot modality of $\mathcal{D}^1$ and $\mathcal{D}^2$ is $b$, and the target modality of $\mathcal{D}^1$ is $c$ because it is unpresented and we hope to infer it. To bind different modalities together, we employ encoders $g^{a}$, $g^{b}$, and $g^{c}$ to project multi-modal data $x^a$, $x^b$, and $x^c$ into latent embeddings $f^a$, $f^b$, and $f^c$, respectively. Our goal is to leverage the mismatched multi-modal datasets $\mathcal{D}^1$ and $\mathcal{D}^2$ and bind all modalities together without bias so that the unpresented modalities can be properly inferred. To achieve this, we first extend Mixup from interpolation to extrapolation.

\subsection{From Interpolation to Extrapolation}
Interpolation through Mixup~\cite{zhang2017mixup} has been an effective data augmentation strategy to enhance generalization ``beyond empirical risk minimization'', which can effectively help CLIP-based multi-modal learning~\cite{oh2024geodesic}. Formally, it conducts a convex combination between two points $f_1$ and $f_2$:
\begin{equation}
    \tilde{f}:=\mathbf{W}\mathbf{F}=
    \begin{bmatrix}
    \lambda & 1-\lambda
    \end{bmatrix}
    \begin{bmatrix}
    f_1 \\
    f_2
    \end{bmatrix},
    \label{eq:interpolation}
\end{equation}
where $\lambda\in\left[0, 1\right]$, $\tilde{f}\in\mathbb{R}^{1\times d}$ is the mixup result of $f_1$ and $f_2$, $\mathbf{W}\in\mathbb{R}^{1\times 2}$ is the weight matrix, and $\mathbf{F}\in\mathbb{R}^{2\times d}$ represents the feature matrix with $d$ dimension. As we can see, Mixup is an interpolation method as it only explores the line segment between two given points, as shown in Figure \ref{fig:extrapolation} (a).

Then, extrapolation can be achieved when $\lambda\in (-\infty, 0]\cup[0, \infty)$ in Eq.~\eqref{eq:interpolation}, which explores outside of the line segment between $f_1$ and $f_2$, as shown in Figure \ref{fig:extrapolation} (b). However, such an extrapolation is still limited to the linear space along the two data points and cannot explore the whole feature space.

Therefore, we consider multi-extrapolation which conducts extrapolation among multiple instances via $\tilde{f}:=\mathbf{W}\mathbf{F}=[w_1, w_2,\ldots,w_n][f_1, f_2,\ldots,f_n]^\top$. As shown in Figure \ref{fig:extrapolation} (c), the extrapolation result can cover the whole feature space if sufficient numbers of data are provided. Further, we extend it to the generation of multiple extrapolation results as
\begin{equation}
    \tilde{\mathbf{F}}:=\mathbf{W}\mathbf{F}=
    \begin{bmatrix}
    w_{1,1} & w_{1,2} & \dots & w_{1,n} \\
    w_{2,1} & w_{2,2} & \dots & w_{2,n} \\
    \vdots & \vdots & \ddots & \vdots \\
    w_{m,1} & w_{m,2} & \dots & w_{m,n}
    \end{bmatrix}
    \begin{bmatrix}
    f_1 \\
    f_2 \\
    \vdots \\
    f_n
    \end{bmatrix},
    \label{eq:extrapolation}
\end{equation}
where $\tilde{\mathbf{F}}\in\mathbb{R}^{m\times d}$, $\mathbf{W}\in\mathbb{R}^{m\times n}$, and $\mathbf{F}\in\mathbb{R}^{n\times d}$. Intuitively, $\mathbf{W}$ acts as a transition matrix to transform from one feature distribution to another. If the relationship between two distributions can be captured via such a transition matrix, we can extrapolate unpresented knowledge based on the presented one. Next, we propose our BrokenBind.

\begin{figure*}
    \centering
    \includegraphics[width=\linewidth]{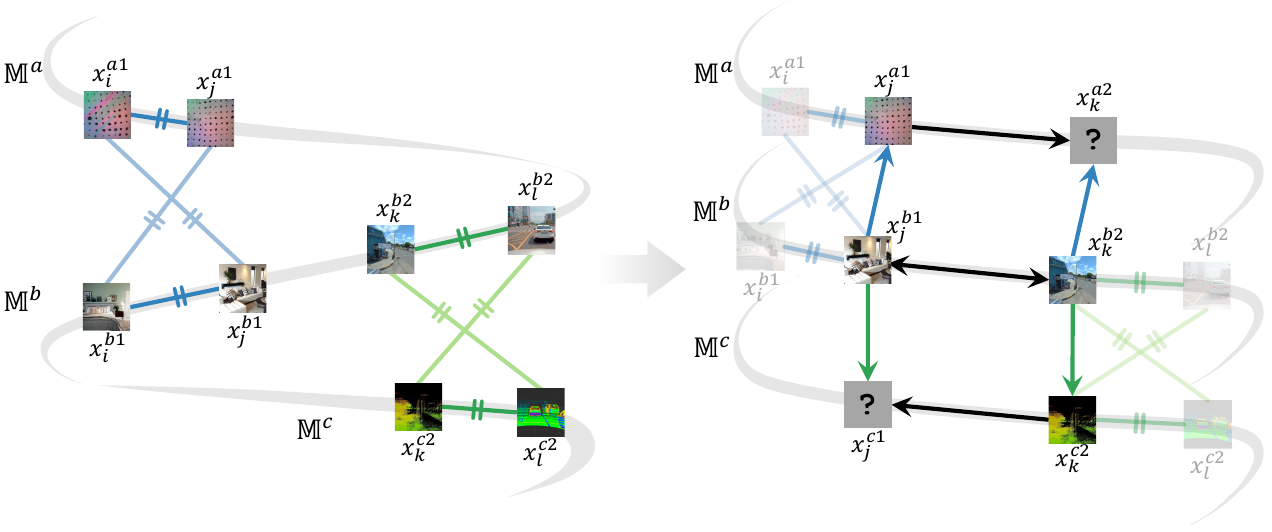}
    \caption{Illustration of cross-modal cross-data consistency. The superscripts denote modality and dataset, subscripts $i$, $j$, $k$, and $l$ are indices, \textit{e.g.}, $x_i^{a1}$ indicates the $i$-th datum from modality $a$ in $\mathcal{D}^1$. Lines with ``$=$'' denote that consistency is applied. The arrows denote the transition relationships from one example to another. Left: Before enforcing consistency, the modality and data structure are inconsistent. Right: After enforcing consistency, a more symmetric structure is presented, benefiting the generalization to novel data and modalities.}
    \label{fig:consistency}
\end{figure*}

\subsection{BrokenBind}
Our learning target for BrokenBind is formulated as:
\begin{equation}
\mathcal{L}=\mathcal{L}_{\text{MOX}}+\mathcal{L}_{\text{CyCLIP}},
\label{eq:bb_framework}
\end{equation}
where the first term is our \emph{Modality Extrapolation} (MOX) for producing and exploring pseudo embeddings of target modality data, and the second term is the CyCLIP training which enforces the cross-data and cross-modal consistency based on the original CLIP framework, as shown in Fig.~\ref{fig:consistency}.

To first conduct MOX, we encode the embedding matrices $\mathbf{F}^{a1}$, $\mathbf{F}^{b1}$, $\mathbf{F}^{b2}$, and $\mathbf{F}^{c2}$ from $\mathcal{D}^1$ and $\mathcal{D}^2$. The extrapolation goal is to produce the pseudo embeddings $\tilde{\mathbf{F}}^{c1}$ and $\tilde{\mathbf{F}}^{a2}$ which is leveraged via:
\begin{equation}
\begin{aligned}
    \mathcal{L}_{\text{MOX}}(c1)\!=\!-\!\frac{1}{2B}\!\sum_{i=1}^B\log\!\left[\frac{\exp(\langle f^{b1}_i, \tilde{f}^{c1}_i\rangle/\tau)}{\sum_{j=1}^B\exp(\langle f^{b1}_i, \tilde{f}^{c1}_j\rangle/\tau)}\right]-\frac{1}{2B}\!\sum_{i=1}^B\log\!\left[\frac{\exp(\langle f^{a1}_i, \tilde{f}^{c1}_i\rangle/\tau)}{\sum_{j=1}^B\exp(\langle f^{a1}_i, \tilde{f}^{c1}_j\rangle/\tau)}\right]\!+\mathcal{R}_{Fro},
\label{eq:mox}
\end{aligned}
\end{equation}
where we denote batch size $B$, temperature scaling factor $\tau$, cosine similarity function $\langle\cdot,\cdot\rangle$. The first two terms contrast $\mathbf{F}^{a1}$ and $\mathbf{F}^{b1}$ from pseudo embeddings $\tilde{\mathbf{F}}^{c1}$, and the last term $\mathcal{R}_{Fro}$ regularizes the pseudo embedding consistency. For another target modality $a$ in $\mathcal{D}^2$, we contrast $\mathbf{F}^{b2}$ and $\mathbf{F}^{c2}$ from $\tilde{\mathbf{F}}^{a2}$ via $\mathcal{L}_{\text{MOX}}(a2)$, so the total MOX loss is $\mathcal{L}_{\text{MOX}}=\mathcal{L}_{\text{MOX}}(c1)+\mathcal{L}_{\text{MOX}}(a2)$. 

In order to obtain $\tilde{\mathbf{F}}^{c1}$ and $\tilde{\mathbf{F}}^{a2}$, we recall Eq.~\eqref{eq:extrapolation} where a transition matrix $\mathbf{W}$ is needed to capture the relationship between different embeddings. As demonstrated in Goel et al.~\cite{goel2022cyclip, jiang2023understanding}, cross-data and cross-modal relationships are two major aspects of constructing multi-modal embedding space. Therefore, we capture them based on the given pivot and target modalities by deriving Eq.~\eqref{eq:extrapolation}:
\begin{align}
    \mathbf{W}^{b\text{-}c}&=\mathbf{F}^{c2}{\mathbf{F}^{b2}}^+, \label{eq:cross_modal} \\
    \mathbf{W}^{2\text{-}1}&=\mathbf{F}^{b1}{\mathbf{F}^{b2}}^+, 
    \label{eq:cross_data}
\end{align}
where ${\mathbf{F}^{b2}}^+$ is the pseudo inverse of $\mathbf{F}^{b2}$~\footnote{We use Moore–Penrose inverse to ensure numerical stability.}, and note that it is only computed on the pivot modality. Eq.~\eqref{eq:cross_modal} is based on the transition process across modality $b$ to $c$, and Eq.~\eqref{eq:cross_data} is based on the transition from datasets $\mathcal{D}^2$ to $\mathcal{D}^1$. Symmetrically, we can capture the relationships from $b$ to $a$ and $\mathcal{D}^1$ to $\mathcal{D}^2$ via transition matrices $\mathbf{W}^{b\text{-}a}$ and $\mathbf{W}^{1\text{-}2}$, respectively.

By applying the cross-modal and cross-data transitions, we can produce pseudo embeddings $\tilde{\mathbf{F}}^{c1}$ via:
\begin{align}
    \tilde{\mathbf{F}}^{c1}_{\text{X-mod}}&=\mathbf{W}^{b\text{-}c}\mathbf{F}^{b1}=\mathbf{F}^{c2}{\mathbf{F}^{b2}}^+\mathbf{F}^{b1}, \label{eq:cross_modal_pseudo} \\
    \tilde{\mathbf{F}}^{c1}_{\text{X-data}}&=\mathbf{W}^{2\text{-}1}\mathbf{F}^{c2}=\mathbf{F}^{b1}{\mathbf{F}^{b2}}^+\mathbf{F}^{c2}. \label{eq:cross_data_pseudo}
\end{align}
Further, our regularization term $\mathcal{R}_{Fro}$ enforces the consistency between the above two pseudo embedding types via:
\begin{equation}
\mathcal{R}_{Fro}(c1)=\|\tilde{\mathbf{F}}^{c1}_{\text{X-mod}}-\tilde{\mathbf{F}}^{c1}_{\text{X-data}}\|_F^2.
\label{eq:frobenius}
\end{equation}
Such a regularization refines $\mathbf{F}^{c2}$ and $\mathbf{F}^{b1}$ around the pivotal embeddings $\mathbf{F}^{b2}$, which helps enhancing the geometric coherence during extrapolation. In this way, the encoder $g^c$ is further updated so that it generalizes well to the unpresented target modality data in $\mathcal{D}^1$. Similarly, regularization $\mathcal{R}_{Fro}(a2)$ can be applied for another target modality $a$ in $\mathcal{D}^2$ which can complete Eq.~\eqref{eq:mox}. During training, the pseudo inverse is always kept fixed, thus it will not cause any computational instabilities.

Intuitively, thanks to the consistency regularization, our MOX extrapolates well-refined pseudo embeddings to complement the missing target modalities. To enhance cross-modal alignment, we contrast between the given non-target modalities and the produced target modalities. As a result, during downstream testing, we can effectively eliminate dataset and modality bias, benefiting generalization to unknown modalities.

Moreover, our BrokenBind leverages CyCLIP~\cite{goel2022cyclip, jiang2023understanding} framework as a backbone to form initial embeddings of exiting modalities, which is composed of CLIP and symmetric losses: $\mathcal{L}_{\text{CyCLIP}}=\mathcal{L}_{\text{CLIP}}+\mathcal{L}_{\text{Sym}}$. In our problem, the CLIP contrastive loss is formed as:
\begin{equation}
\begin{aligned}
    \!\!\!\!\!\!\mathcal{L}_{\text{CLIP}}(a1)\!=\!&-\frac{1}{3B}\!\sum_{i=1}^B\log\!\left[\frac{\exp(\langle f^{a1}_i, f^{b1}_i\rangle/\tau)}{\sum_{j=1}^B\exp(\langle f^{a1}_i, f^{b1}_j\rangle/\tau)}\right]\\
    &+\frac{1}{3B}\!\sum_{i=1}^B\log\!\left[\sum_{k=1}^B\exp(\langle f^{a1}_i, f^{b2}_k\rangle/\tau)\right]\\
    &+\frac{1}{3B}\!\sum_{i=1}^B\log\!\left[\sum_{l=1}^B\exp(\langle f^{a1}_i, f^{c2}_l\rangle/\tau)\right],\!\!\!
\end{aligned}
\label{eq:one_side_clip}
\end{equation}
The loss function $\mathcal{L}_{\text{CLIP}}(a1)$ aims to pull the positive pairs between modality $a$ and modality $b$ in dataset $\mathcal{D}^1$ together, meanwhile all the unpaired examples from the rest modalities and datasets are considered negative pairs, thus pushed further. Then, similar losses from each modality and each dataset are combined as our total CLIP loss $\mathcal{L}_{\text{CLIP}}=\mathcal{L}_{\text{CLIP}}(a1)+\mathcal{L}_{\text{CLIP}}(b1)+\mathcal{L}_{\text{CLIP}}(b2)+\mathcal{L}_{\text{CLIP}}(c2)$.

\begin{algorithm}[t]
\caption{BrokenBind}
\label{alg:brokenbind}
\begin{algorithmic}[1]
\REQUIRE \text{ }\\
Learning model: \texttt{model}() parameterized by $\theta$ \\
Loss functions: \texttt{mox}() realizes Eq.~\eqref{eq:mox}, \texttt{contrastive}() realizes Eq.~\eqref{eq:one_side_clip}, and \texttt{symmetry}() realizes Eqs.~\eqref{eq:cross_modal_consistency} and~\eqref{eq:cross_data_consistency}
\ENSURE Dataloaders \texttt{loader1} and \texttt{loader2}
\FOR{$t=1$ to $T$}
\STATE Sample ($x^{a1}$, $x^{b1}$) from \texttt{loader1}, ($x^{b2}$ and $x^{c2}$) from \texttt{loader2};
\STATE Encode $f^{a1}$, $f^{b1}$, $f^{b2}$, and $f^{c2}$ via \texttt{model}($x^{a1}, x^{b1}, x^{b2}, x^{c2}$);
\STATE Compute ${f^{b2}}^{+}$ and ${f^{b1}}^{+}$ through pseudo inverse, and stop gradient;
\STATE Compute $\tilde{f}^{c1}_{\text{X-mod}}$, $\tilde{f}^{c1}_{\text{X-data}}$, $\tilde{f}^{a2}_{\text{X-mod}}$, and $\tilde{f}^{a2}_{\text{X-data}}$ via in Eqs.~\eqref{eq:cross_modal_pseudo} and~\eqref{eq:cross_data_pseudo};
\STATE $\mathcal{L}_{\text{MOX}}=$\texttt{mox}$(f^{a1}, f^{b1}, \tilde{f}^{c1}_{\text{X-data}}, \tilde{f}^{a2}_{\text{X-data}} f^{b2}, f^{c2})$;
\STATE $\mathcal{L}_{\text{CLIP}}=$\texttt{contrastive}$(f^{a1}, f^{b1}, f^{b2}, f^{c2})$;
\STATE $\mathcal{L}_{\text{Sym}}=$\texttt{symmetry}$(f^{a1}, f^{b1}, f^{b2}, f^{c2})$;
\STATE $\mathcal{L}=\mathcal{L}_{\text{MOX}}+\mathcal{L}_{\text{CLIP}}+\mathcal{L}_{\text{Sym}}$
\STATE Backpropagate $\mathcal{L}$ and optimize $\theta$.
\ENDFOR
\RETURN \texttt{model}()
\end{algorithmic}
\end{algorithm}

As for symmetric loss, CyCLIP proposed cross-modal and cross-data symmetry assumptions where the embedding structure across different modalities or different data examples should be symmetric, which can be achieved by minimizing the following loss:
\begin{align}
    \!\!\!\!\!\mathcal{L}_{\text{X-mod}}(a1, b1)=&\frac{1}{B}\sum_{i=1}^B\sum_{j=1}^B(\langle f^{a1}_i, f^{b1}_j \rangle\! - \!\langle f^{a1}_j, f^{b1}_i \rangle)^2, \!\label{eq:cross_modal_consistency} \\
    \!\!\!\!\!\mathcal{L}_{\text{X-data}}(a1, b1)=&\frac{1}{B}\sum_{i=1}^B\sum_{j=1}^B(\langle f^{a1}_i, f^{a1}_j \rangle\! - \!\langle f^{b1}_i, f^{b1}_j \rangle)^2. \!\label{eq:cross_data_consistency}
\end{align}
As shown in the left of Figure \ref{fig:consistency}, given $(x^{a1}_i, x^{b1}_i)$ and $(x^{a1}_j, x^{b1}_j)$ from $\mathcal{D}^1$, the distance between $x^{a1}_i$ and $x^{b1}_j$ should be similar to the distance between $x^{a1}_i$ and $x^{b1}_j$. Meanwhile, the distance between $x^{a1}_i$ and $x^{a1}_j$ should be similar to the distance between $x^{b1}_i$ and $x^{b1}_j$. Similarly, we apply Eqs.~\eqref{eq:cross_modal_consistency} and~\eqref{eq:cross_data_consistency} to dataset $\mathcal{D}^2$ to obtain our total symmetric loss $\mathcal{L}_{\text{Sym}}=\mathcal{L}_{\text{X-mod}}(a1, b1)+\mathcal{L}_{\text{X-data}}(a1, b1)+\mathcal{L}_{\text{X-mod}}(b2, c2)+\mathcal{L}_{\text{X-data}}(b2, c2)$. 

As shown in CyCLIP studies~\cite{goel2022cyclip, jiang2023understanding}, enforcing symmetry across modality and dataset can enhance the geometric coherence of the embedding space, which is vital for downstream generalization to novel image-text pairs. As illustrated by Figure \ref{fig:consistency}, both different modality spaces and different data pairs are more symmetric in the right figure compared to the left one. Replying to the consistency across modalities and data examples, our modality extrapolation leads to improved effectiveness, as shown in our ablation studies in Sec.~\ref{sec:experiments}. We summarize the process of BrokenBind in Algorithm \ref{alg:brokenbind}. Next, we conduct experiments to show that our method can effectively generalize to unseen modalities and achieve significant improvement on emergent alignment through modality binding.

\section{Experiments}
\label{sec:experiments}
In this section, we conduct extensive experiments to show the effectiveness of BrokenBind on binding modalities from mismatched datasets. First, we describe basic implementation details, such as datasets, models, experimental setup, and baseline methods. Then, we provide quantitative comparisons between BrokenBind and other well-known models and methods. Further, we quantitatively show the generalization improvement thanks to our modality extrapolation. Finally, we study the ablation results of learning modules, hyper-parameters, and training efficiency.

\subsection{Implementation Details}
\paragraph{Experimental setup:}
To train the models, we conduct LoRA fine-tuning~\cite{hu2021lora} and freeze the backbone encoders. Moreover, to avoid distorting features by fine-tuning~\cite{kumar2022fine}, we first conduct linear probing to train the projectors, then load the adapted projectors to conduct LoRA fine-tuning. The trainable parameters are optimized by AdamW optimizer with an initial learning rate of $5e-4$ with weight decay $0.2$ for $50$ epochs. We conduct consistency pre-training in the first $25$ epochs and enable modality extrapolation for the rest $25$ epochs. Moreover, we consider two BrokenBind settings where modalities from different numbers of datasets are bound together, as shown in Table \ref{table:binding_setting}. Intuitively, when we hope to infer the tactile modality ``ta'' in ObjF-real by leveraging ObjF-1.0, then vision ``vi'' acts as a pivot modality. For simplicity, we denote a dataset flow as ``ObjF-1.0$\rightarrow$ObjF-real'' and modality flow as ``te$\rightarrow$vi$\rightarrow$ta''. Similarly, for the setting with 3 datasets, we have dataset flow and modality flow as ``ObjF-1.0$\rightarrow$AVE$\rightarrow$ObjF-real'' and ``te$\rightarrow$vi$\rightarrow$ta'', respectively. For evaluation, we based on the modality flow and measure the connection between the beginning modality and target modality using mean average precision (mAP) \footnote{We use mAP instead of Recall because the Recall of both ImageBind and LanguageBind are low.}. For example, given ``te$\rightarrow$vi$\rightarrow$ta'', we report the test mAP of using ``te'' to retrieve ``ta''.

\setlength{\intextsep}{3pt}
\setlength{\columnsep}{8pt}
\begin{wraptable}{r}{0.4\textwidth}
\centering
\vspace{-3mm}
\setlength{\tabcolsep}{10pt}
\caption{Our experimental settings using different dataset numbers. The checkmark \ding{51} denotes that a modality is observable or cross \ding{55} otherwise. The circled cross \textcircled{\ding{55}} indicates the target modality to be inferred.}
\begin{tabular}{l|l|ccc}
\toprule[1.2pt]
\textbf{\#  } & \textbf{Dataset} & \textbf{te} & \textbf{vi} & \textbf{ta} \\ \midrule
\multirow{2}{*}{2} & ObjF-1.0 & \textcircled{\ding{55}} & \ding{51} & \ding{51} \\
& ObjF-real & \ding{51} & \ding{51} & \textcircled{\ding{55}} \\
\midrule
\multirow{3}{*}{3} & ObjF-1.0 & \ding{55} & \ding{51} & \ding{51} \\
& AVE & \ding{51} & \ding{51} & \ding{55} \\
& ObjF-real & \ding{51} & \ding{55} & \textcircled{\ding{55}} \\
\bottomrule[1.2pt]
\end{tabular}
\label{table:binding_setting}
\vspace{-5mm}
\end{wraptable}
\paragraph{Datasets:}
We consider 6 modalities in total, namely vision (``vi''), text (``te''), audio (``au''), tactile sensing (``ta''), depth (``de''), and thermal (``th''). Based on these modalities, we leverages several well-known multi-modal datasets, namely MSRVTT~\cite{xu2016msr}, AVE~\cite{tian2018audio}, ObjF (including ``real'' and ``1.0'')~\cite{gao2021ObjectFolder, gao2022ObjectFolderV2}, NYU~\cite{silberman2012indoor}, SUN~\cite{song2015sun}, LLVIP~\cite{jia2021llvip}, and FLIR~\cite{flir_adas_dataset}. In order to evaluate the performance of our method, we choose two datasets that contain a target modality and make the target modality of one dataset invisible during training, then we can test the performance of the target modality. For example, considering depth as the target modality, we choose NYU-D and SUN-D datasets and hide the depth data from each dataset. Then, we use vision as the pivot modality to infer depth data.

\paragraph{Models:}
In our experiments, we choose the corresponding encoders from ImageBind~\cite{girdhar2023imagebind}, LanguageBind~\cite{zhu2023languagebind}, Video-CLIP~\cite{xu2021videoclip}, CLAP~\cite{elizalde2023clap}, and TVL~\cite{fu2024touch}. ImageBind covers vision, text, audio, thermal, depth, and IMU modalities, LanguageBind covers vision, text, audio, thermal, and depth modalities, Video-CLIP covers vision, and text modalities, CLAP covers vision and audio, and TVL covers vision, text, and tactile modalities. For simplicity, we denote ImageBind as ``IB'', LanguageBind as ``LB'', Video-CLIP as ``VCLIP'', and our BrokenBind as ``BB''.

\paragraph{Baseline methods:}
For performance comparison, we first use the backbone models mentioned above as the baseline methods. Moreover, we directly train them under the BrokenBind setting to form fine-tuned baselines (``+FT'').

\begin{table*}[t]
\renewcommand{\arraystretch}{1.0} 
\caption{Performance under various dataset-modality settings. Models include IB and LB, and modalities include au, vi, te, de, and th.}
\setlength{\tabcolsep}{1.6pt}
\begin{tabular}{@{}lcccccccccccc@{}}
\toprule[1.2pt]
\multirow{2}{*}{Method} & \multicolumn{2}{c}{AVE $\rightarrow$ MSRVTT} & \multicolumn{2}{c}{MSRVTT $\rightarrow$ AVE} & \multicolumn{2}{c}{SUN $\rightarrow$ NYU} & \multicolumn{2}{c}{NYU $\rightarrow$ SUN} & \multicolumn{2}{c}{FLIR $\rightarrow$ LLVIP} & \multicolumn{2}{c}{LLVIP $\rightarrow$ FLIR} \\
\cmidrule(lr){2-3} \cmidrule(lr){4-5} \cmidrule(lr){6-7} \cmidrule(lr){8-9} \cmidrule(lr){10-11} \cmidrule(lr){12-13}
& \scriptsize au$\rightarrow$vi$\rightarrow$te & \scriptsize te$\rightarrow$vi$\rightarrow$au & \scriptsize au$\rightarrow$vi$\rightarrow$te & \scriptsize te$\rightarrow$vi$\rightarrow$au & \scriptsize de$\rightarrow$vi$\rightarrow$te & \scriptsize te$\rightarrow$vi$\rightarrow$de & \scriptsize de$\rightarrow$vi$\rightarrow$te & \scriptsize te$\rightarrow$vi$\rightarrow$de & \scriptsize th$\rightarrow$vi$\rightarrow$te & \scriptsize te$\rightarrow$vi$\rightarrow$th & \scriptsize th$\rightarrow$vi$\rightarrow$te & \scriptsize te$\rightarrow$vi$\rightarrow$th \\
\midrule
IB                & 11.4 & 9.2  & 8.1  & 5.3  & 13.5 & 10.0   & 7.5  & 3.9  & 15.8 & 13.5 & 14.1 & 11.6 \\
IB+FT           & 12.4 & 11.0   & 9.3  & 5.7  & 15.1 & 12.4 & 13.2 & 12.5 & 21.2 & 18.0   & 19.6 & 18.9 \\
IB+BB   & \textbf{35.3} & \textbf{34.1} & \textbf{27.8} & \textbf{25.6} & \textbf{38.7} & \textbf{35.5} & \textbf{31.2} & \textbf{33.1} & \textbf{42.5} & \textbf{36.9} & \textbf{40.2} & \textbf{37.6} \\
\midrule
LB             & 13.4 & 10.1 & 11.2 & 8.9  & 12.6 & 11.5 & 8.9  & 4.1  & 14.7 & 14.4 & 13.8 & 12.5 \\
LB+FT        & 15.3 & 14.1 & 13.6 & 10.6 & 15.2 & 14.0   & 13.6 & 13.2 & 23.4 & 21.3 & 21.3 & 19.9 \\
LB+BB & \textbf{32.5} & \textbf{33.4} & \textbf{29.2} & \textbf{24.8} & \textbf{36.8} & \textbf{36.1} & \textbf{32.1} & \textbf{29.4} & \textbf{45.2} & \textbf{43.0}   & \textbf{39.4} & \textbf{36.8} \\
\bottomrule[1.2pt]
\end{tabular}
\label{tab:comparison_1}
\end{table*}

\begin{table}[t]
\centering
\noindent
\begin{minipage}{0.49\textwidth}
\centering
\renewcommand{\arraystretch}{1.0} 
\setlength{\tabcolsep}{2.5pt} 
\caption{Performance under various dataset-modality settings. Models using TVL backbone, and modalities include vi, te, and ta.}
\label{tab:comparison_2}
\begin{tabular}{@{}lcccc@{}}
\toprule[1.2pt] 
\multirow{2}{*}{Method} & \multicolumn{2}{c}{ObjF-1.0 $\rightarrow$ ObjF-real} & \multicolumn{2}{c}{ObjF-real $\rightarrow$ ObjF-1.0} \\
\cmidrule(lr){2-3} \cmidrule(lr){4-5}
& \scriptsize ta$\rightarrow$vi$\rightarrow$te & \scriptsize te$\rightarrow$vi$\rightarrow$ta & \scriptsize ta$\rightarrow$vi$\rightarrow$te & \scriptsize te$\rightarrow$vi$\rightarrow$ta \\
\midrule
TVL              & 15.2 & 14.2 & 12.7 & 9.6 \\
TVL+FT           & 18.5 & 16.8 & 16.7 & 14.1 \\
TVL+BB           & \textbf{28.5} & \textbf{26.0} & \textbf{29.3} & \textbf{25.4} \\
\bottomrule[1.2pt]
\end{tabular}
\end{minipage}
\hfill
\begin{minipage}{0.49\textwidth}
\centering
\renewcommand{\arraystretch}{1.0} 
\setlength{\tabcolsep}{1.8pt} 
\caption{Performance under various settings. Model using VCLIP combined with CLAP, and modalities include vi, te, and au.}
\begin{tabular}{@{}lcccc@{}}
\toprule[1.2pt] 
\multirow{2}{*}{Method} & \multicolumn{2}{c}{AVE $\rightarrow$ MSRVTT} & \multicolumn{2}{c}{MSRVTT $\rightarrow$ AVE} \\
\cmidrule(lr){2-3} \cmidrule(lr){4-5}
& \scriptsize au$\rightarrow$vi$\rightarrow$te & \scriptsize te$\rightarrow$vi$\rightarrow$au & \scriptsize au$\rightarrow$vi$\rightarrow$te & \scriptsize te$\rightarrow$vi$\rightarrow$au \\
\midrule
VCLIP+CLAP              & 15.8 & 13.6 & 13.2 & 11.1 \\
VCLIP+CLAP+FT           & 17.6 & 15.4 & 17.2 & 13.6 \\
VCLIP+CLAP+BB           & \textbf{36.4} & \textbf{37.1} & \textbf{33.1} & \textbf{28.7} \\
\bottomrule[1.2pt]
\end{tabular}
\label{tab:comparison_3}
\end{minipage}
\end{table}

\subsection{Quantitative Comparisons}

\paragraph{Binding through Two Datasets:}
First, we consider binding modalities from two datasets. The results under various modalities and dataset settings using different backbone encoders are shown in Tables \ref{tab:comparison_1}, \ref{tab:comparison_2}, and \ref{tab:comparison_3}. We observe a very limited performance of the naively employed encoders because the chosen datasets are not seen during their pre-training. Moreover, we find further fine-tuning on the observed modalities in each dataset. For example, for ``AVE $\rightarrow$ MSRVTT'' with ``au $\rightarrow$ vi $\rightarrow$ te'' setting, we train modalities au and vi from the AVE dataset via contrastive learning, and the same for modalities vi and te from MSRVTT. Then, the performance is measured by the au-te pairs from the test set of MSRVTT. Surprisingly, such a fine-tuning strategy brings very limited performance improvement and is even harmful in some scenarios. This is because the mismatched distributions between different datasets hinder the modality binding, which explains why existing studies show limited emergent alignment performances. Fortunately, our BrokenBind can bridge the dataset gap through modality exploration, ensuring that contrastive learning is conducted without significant bias, thus benefiting the binding performance across datasets. As we can see, BrokenBind significantly surpasses other baseline methods under various settings and backbone models.

\begin{wraptable}{r}{0.55\textwidth}
\centering
\vspace{-3mm}
\renewcommand{\arraystretch}{1.0} 
\setlength{\tabcolsep}{1pt} 
\caption{Performance under various settings using three datasets. Models using IB and LB, and modalities include vi, te, and de.}
\begin{tabular}{@{}lcccc@{}}
\toprule[1.2pt] 
\multirow{2}{*}{Method} & \multicolumn{2}{c}{SUN $\rightarrow$ MSRVTT $\rightarrow$ NYU} & \multicolumn{2}{c}{NYU $\rightarrow$ MSRVTT $\rightarrow$ SUN} \\
\cmidrule(lr){2-3} \cmidrule(lr){4-5} &
\scriptsize de$\rightarrow$vi$\rightarrow$te & \scriptsize te$\rightarrow$vi$\rightarrow$de & \scriptsize de$\rightarrow$vi$\rightarrow$te & \scriptsize te$\rightarrow$vi$\rightarrow$de  \\
\midrule
IB              & 13.5 & 10.0 & 7.5 & 3.9 \\
IB+FT           & 12.6 & 9.8 & 8.1 & 5.6 \\
IB+BB           & \textbf{27.2} & \textbf{25.4} & \textbf{26.8} & \textbf{22.5} \\
\midrule
LB              & 12.6 & 11.5 & 8.9 & 4.1 \\
LB+FT           & 13.4 & 12.1 & 10.6 & 6.7 \\
LB+BB           & \textbf{25.6} & \textbf{24.3} & \textbf{25.7} & \textbf{24.6} \\
\bottomrule[1.2pt]
\end{tabular}
\label{tab:comparison_multipledatasets_1}
\end{wraptable}
\paragraph{Binding through Three Datasets:}
Then, we consider an intensified scenario where we hope to bind modalities across three datasets. Such a scenario occurs when the two modalities to be bound are from two datasets, and the two datasets do not have any shared modalities. Thus, we have to acquire an additional dataset that has shared modalities with both two datasets. To tackle this problem, we can break down and solve the standard two-dataset problem. For example, for ``ObjF-1.0 $\rightarrow$ AVE $\rightarrow$ ObjF-real'' with ``ta $\rightarrow$ vi $\rightarrow$ te'', we first conduct two-dataset BrokenBind using ObjF-real and AVE to explore the modality vi. Though vi is not the target modality, it can act as a pivot modality that is shared by ObjF-1.0. Hence, we can combine ObjF-1.0 and ObjF-real to complete modality binding between ta and te. To evaluate the experimental performance, we show the results in Tables \ref{tab:comparison_multipledatasets_1} and \ref{tab:comparison_multipledatasets_2}. In this challenging setting, we can see that fine-tuning is even harmful to binding two modalities across three datasets because the gap between datasets is further intensified compared to two-dataset scenarios. However, BrokenBind still shows effective performance compared to both baselines, which justifies the binding capability of our method under extreme settings.

\begin{wraptable}{r}{0.55\textwidth}
\centering
\vspace{-3mm}
\renewcommand{\arraystretch}{1.0} 
\setlength{\tabcolsep}{0.5pt} 
\caption{Performance under various settings using three datasets. Model using TVL, and modalities include vi, te, and ta.}
\begin{tabular}{@{}lcccc@{}}
\toprule[1.2pt] 
\multirow{2}{*}{Method} & \multicolumn{2}{c}{ObjF-1.0$\!\rightarrow\!$AVE$\!\rightarrow\!$ObjF-real} & \multicolumn{2}{c}{ObjF-real$\!\rightarrow\!$AVE$\!\rightarrow\!$ObjF-1.0} \\
\cmidrule(lr){2-3} \cmidrule(lr){4-5}
& \scriptsize ta$\!\rightarrow\!$vi$\!\rightarrow\!$te & \scriptsize te$\!\rightarrow\!$vi$\!\rightarrow\!$ta & \scriptsize ta$\!\rightarrow\!$vi$\!\rightarrow\!$te & \scriptsize te$\!\rightarrow\!$vi$\!\rightarrow\!$ta \\
\midrule
TVL              & 15.2 & 14.2 & 12.7 & 9.6 \\
TVL+FT           & 15.5 & 13.8 & 11.1 & 9.0 \\
TVL+BB           & \textbf{21.2} & \textbf{19.5} & \textbf{20.1} & \textbf{18.4} \\
\bottomrule[1.2pt]
\end{tabular}
\label{tab:comparison_multipledatasets_2}
\end{wraptable}
\paragraph{Binding Emergent Modalities:}
Further, we study the emergent alignment problem where the two modalities are not directly aligned together during pre-training. For example, LanguageBind only binds various modalities to te modality, hence, de and vi are not directly bound. Practically, the reason why two modalities are not bound together is mostly because they are rare and there is no existing dataset. Therefore, we consider such a realistic scenario and tackle it using BrokenBind, as shown in Tables \ref{tab:comparison_emergent_1} and \ref{tab:comparison_emergent_2}. We can see that fine-tuning is still ineffective in improving the performance of binding emergent modalities. Similarly, the superiority of BrokenBind can be justified by the large performance boost on various settings using different backbone encoders.

\begin{table}[t]
\centering
\noindent
\begin{minipage}{0.45\textwidth}
\renewcommand{\arraystretch}{1.0} 
\setlength{\tabcolsep}{6pt} 
\caption{Performance of binding emergent modalities across datasets. Model using LB, and modalities include vi, te, and de.}
\begin{tabular}{@{}lcccc@{}}
\toprule[1.2pt] 
\multirow{2}{*}{Method} & \multicolumn{2}{c}{SUN $\rightarrow$ NYU} & \multicolumn{2}{c}{NYU $\rightarrow$ SUN} \\
\cmidrule(lr){2-3} \cmidrule(lr){4-5}
& \scriptsize de$\rightarrow$te$\rightarrow$vi & \scriptsize vi$\rightarrow$te$\rightarrow$de & 
\scriptsize de$\rightarrow$te$\rightarrow$vi & \scriptsize vi$\rightarrow$te$\rightarrow$de \\
\midrule
LB              & 9.2 & 8.4 & 6.4 & 7.7 \\
LB+FT           & 10.2 & 9.6 & 8.9 & 9.2 \\
LB+BB           & \textbf{19.3} & \textbf{18.5} & \textbf{16.5} & \textbf{16.0} \\
\bottomrule[1.2pt]
\end{tabular}
\label{tab:comparison_emergent_1}
\end{minipage}
\hfill
\begin{minipage}{0.53\textwidth}
\centering
\renewcommand{\arraystretch}{1.0} 
\setlength{\tabcolsep}{4.8pt} 
\caption{Performance of binding emergent modalities across datasets. Model using VLIP and CLAP, and modalities include vi, te, and au.}
\begin{tabular}{@{}lcccc@{}}
\toprule[1.2pt] 
\multirow{2}{*}{Method} & \multicolumn{2}{c}{AVE $\rightarrow$ MSRVTT} & \multicolumn{2}{c}{MSRVTT $\rightarrow$ AVE} \\
\cmidrule(lr){2-3} \cmidrule(lr){4-5}
& \scriptsize au$\rightarrow$te$\rightarrow$vi & \scriptsize vi$\rightarrow$te$\rightarrow$au & 
\scriptsize au$\rightarrow$te$\rightarrow$vi & \scriptsize vi$\rightarrow$te$\rightarrow$au \\
\midrule
VCLIP+CLAP              & 5.1 & 6.2 & 4.2 & 3.8 \\
VCLIP+CLAP+FT           & 8.4 & 9.6 & 7.0 & 6.2 \\
VCLIP+CLAP+BB           & \textbf{17.6} & \textbf{17.4} & \textbf{16.8} & \textbf{18.1} \\
\bottomrule[1.2pt]
\end{tabular}
\label{tab:comparison_emergent_2}
\end{minipage}
\end{table}

\subsection{Qualitative Analyses}

\begin{wrapfigure}{r}{0.55\textwidth}
    \centering
    \includegraphics[width=0.55\textwidth]{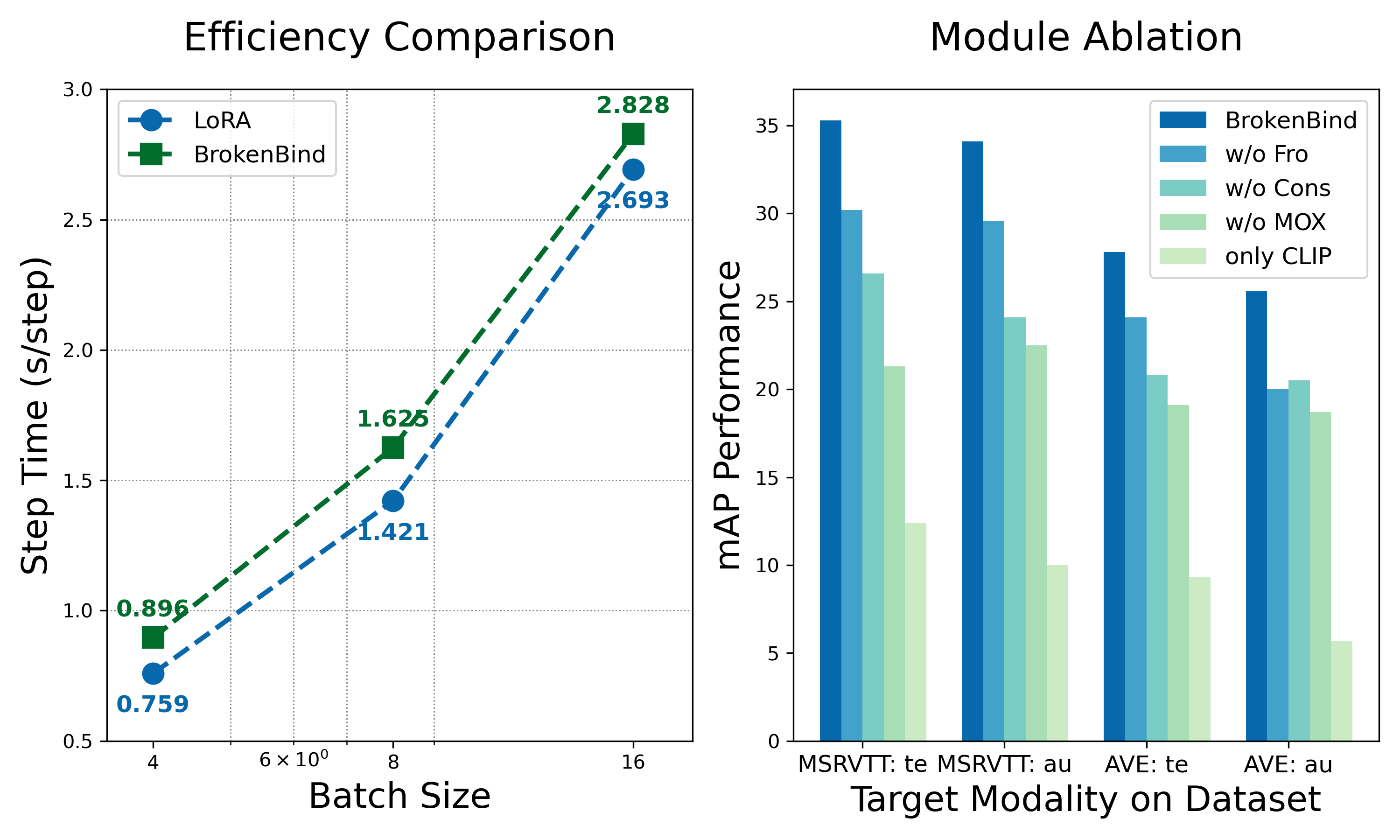}
    \caption{Left: Efficiency study between BrokenBind and LoRA fine-tuning. Right: Ablation study on different module settings.}
    \label{fig:ablation}
\end{wrapfigure}

\paragraph{Computational Efficiency:}
To study the computational efficiency of BrokenBind, we compare the optimization speed of BrokenBind to plain LoRA fine-tuning, as shown in the left of Figure \ref{fig:ablation}. Compared to LoRA fine-tuning, our method adds extra loss computation, such as consistency training and modality extrapolation, which is backpropagated along with the contrastive loss. Hence, we can see that the computation step time of our method is not significantly higher than LoRA fine-tuning.

\paragraph{Ablation Study:}
Then, to study the effectiveness of each module in BrokenBind, we remove each of them to form different settings. Particularly, we have the full ``BrokenBind'' setting, removing the Frobenius regularization ``w/o Fro'', removing the consistency training ``w/o Cons'', and removing the modality extrapolation ``w/o MOX''. Moreover, we consider the plain CLIP training ``only CLIP'' which is the same for plain fine-tuning. The result is shown in the right Figure \ref{fig:ablation}. We can see that BrokenBind achieves the best performance among all module settings, which justifies the effectiveness of our methodology design. Additionally, we find that modality extrapolation has the biggest influence in contributing to the final performance. Secondly, consistency training helps regularize the geometric symmetry which is also important for generalization. When removing all the modules and only training with CLIP contrastive learning, the performance drops to the worst, because such a strategy is sub-optimal under dataset mismatch, further harming the binding performance.

\begin{wraptable}{r}{0.5\textwidth}
\centering
\vspace{-3mm}
\renewcommand{\arraystretch}{1.0} 
\setlength{\tabcolsep}{1pt} 
\caption{Performance study of linear probing.}
\begin{tabular}{@{}lcccc@{}}
\toprule[1.2pt] 
\multirow{2}{*}{Method} & \multicolumn{2}{c}{AVE $\rightarrow$ MSRVTT} & \multicolumn{2}{c}{MSRVTT $\rightarrow$ AVE} \\
\cmidrule(lr){2-3} \cmidrule(lr){4-5}
& \scriptsize au$\rightarrow$vi$\rightarrow$te & \scriptsize te$\rightarrow$vi$\rightarrow$au & \scriptsize au$\rightarrow$vi$\rightarrow$te & \ te$\rightarrow$vi$\rightarrow$au \\
\midrule
Linear Prob.              & 14.3 & 12.1 & 11.1 & 8.6 \\
LoRA           & 23.2 & 18.3 & 20.1 & 17.4 \\
Linear Prob.\&LoRA           & \textbf{35.3} & \textbf{34.1} & \textbf{27.8} & \textbf{25.6} \\
\bottomrule[1.2pt]
\end{tabular}
\label{tab:linear_probing}
\end{wraptable}
\paragraph{Effectiveness of Linear Probing}
Further, we study the performance of linear probing. As mentioned in Kumar et al.~\cite{kumar2022fine}, directly conducting fine-tuning could distort the feature embeddings. Hence, we first conduct linear probing to adapt the projectors, then further fine-tune via LoRA. To study the effectiveness of linear probing, we show the result in Table \ref{tab:linear_probing}. Specifically, we consider three settings: only conduct linear probing, only conduct LoRA fine-tuning, and first use linear probing then fine-tune with LoRA. We can see that using linear probing before LoRA fine-tuning achieves the best performance. However, directly conducting LoRA fine-tuning shows limited performance improvement, which might be caused by the embedding distortion that hinders generalization. Moreover, only linear probing cannot show effective performance because the projection layer is too shallow, failing to capture the essential knowledge of downstream datasets.

\begin{wrapfigure}{r}{0.65\textwidth}
    \centering
\vspace{-3mm}
    \includegraphics[width=\linewidth]{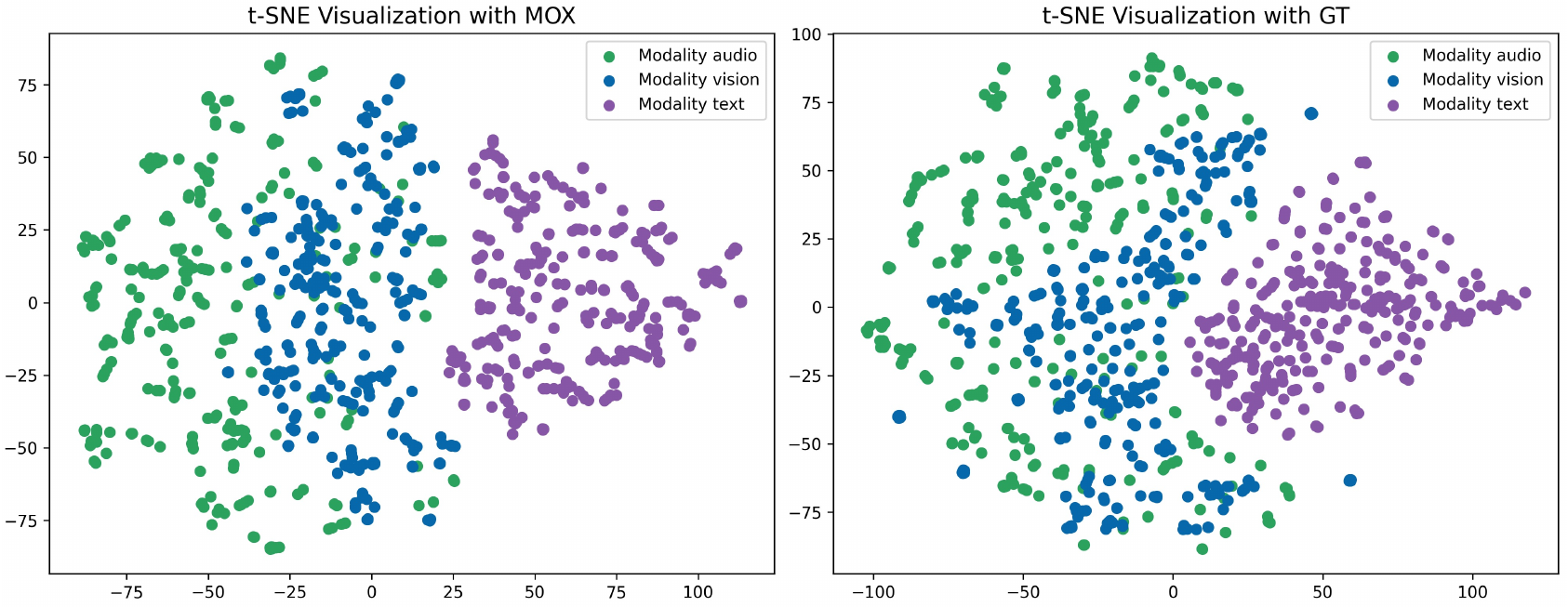}
    \caption{t-SNE visualization of embeddings. Left: modality extrapolation (MOX) on text. Right: ground truth (GT) embeddings.}
    \label{fig:tsne}
\end{wrapfigure}

\paragraph{t-SNE Visualization:}
To further qualitatively study the effectiveness of our modality extrapolation, we use t-SNE visualization to show the extrapolation result compared to the ground truth embeddings in Figure \ref{fig:tsne}. Here we use text as the target modality and follow Eq. \eqref{eq:cross_data_pseudo} to produce pseudo text embeddings as shown by the purple scatters. We can see that the pseudo embeddings are distributed at the same place compared to the ground truth. Therefore, we justify that our extrapolation can effectively compensate for the missing modality to help learn contrastive representations from the two other modalities.

\section{Conclusion and Discussion}
In this paper, we study a realistic modality binding problem where two modalities to be bound are not available in the same dataset, which requires us to leverage multiple datasets. However, due to the significant gap that exists in different datasets, the modality binding performance is largely biased. Therefore, to solve this problem, we propose BrokenBind which captures the relationships across datasets and modalities to conduct modality extrapolation. The extrapolation process can effectively produce pseudo embeddings that compensate for the missing modalities in each dataset, allowing us to conduct contrastive representation learning without significant bias. Hence, we can successfully bind two modalities together and achieve improved generalization performance. Through extensive experiments, we carefully justified the performance of our BrokenBind under various settings, including the intensified scenario where modality binding is conducted across three datasets. Moreover, we provide intuitive performance analyses to quantitatively evaluate our method.

Moreover, there are a few potential limitations of our work. First, we are based on several popular pre-trained binding models, which have very limited performance in some cases, such as tactile sensing with text. The reason is that the performance from different modalities is highly imbalanced due to their commonness. Therefore, the performance of BrokenBind under rare modalities is still limited. In the future, we hope to explore such an imbalance problem and hope to leverage the common modalities to enhance the binding performance of rare modalities.

\bibliographystyle{unsrt}
\bibliography{references}  






\appendix
\clearpage

\begin{center}
 \rule{6.50in}{1.2pt}\\
 {\Large\bf Supplementary Material for ``\textsc{BrokenBind}: \\ \vspace{0.06in} Universal Modality Exploration beyond Dataset Boundaries''}
 \rule{6.50in}{1.2pt}
\end{center}

\vskip 0.3in

In this supplementary material, we provide additional discussions and experimental results to comprehensively demonstrate the effectiveness of BrokenBind. First, we discuss more detailed implementation details. Then, we carefully demonstrate the extension of BrokenBind from two-dataset binding to three-dataset binding, and even multiple-dataset binding ($>3$). Further, we conduct additional empirical analyses, including a comparison to methods that connect modalities, visualization, and ablation on additional datasets, and zero-shot improvement for emergent alignments.

\section{Additional Details}
Our experiments can be conducted on a single NVIDIA A100/H100 or 4$\times$NVIDIA 4090 GPU. In our experiments, we use batch size 16 which contains all considered datasets. For example, if we aim to bind from two datasets, then we equally sample 8 examples from each dataset in one batch. Similarly, each dataset is sampled with $\sim$5 examples if we bind from three datasets.

To transform the multi-modal data, we follow ImageBind~\cite{girdhar2023imagebind}. For vision data, we conduct standard RGB and RGBT representations, where the video data has 2-frame clips with patch size $2\times16\times16$ ($T\times H\times W$). For thermal data, we use single-channel images. For depth data, we use disparity representations. For audio data, we use the waveform and transform it into mel-spectrograms~\cite{gong2021ast}. For tactile data, we follow TVL~\cite{fu2024touch} to perform normalization and background subtraction on the 224$\times$224 sized input.

To conduct basic contrastive training, we first extract feature embeddings using various modality encoders. Then, we use a temperature scale on top of the normalization layer. Specifically, for vision data, text data, audio data, depth data, thermal data, and tactile data, the scaling factors are set to $1.0$, $1.0$, $20.0$, $5.0$, $10.0$, and $10.0$, respectively.

\section{Extension to Three Datasets}
To extend our BrokenBind to multiple datasets, here we provide a simple three-dataset example. Specifically, we are given $\mathcal{D}^1=\{x_i^{a1}\}_{i=1}^l$, $\mathcal{D}^2=\{x_j^{a2}, x_j^{b2}\}_{j=1}^m$, and $\mathcal{D}^3=\{x_k^{b3}, x_k^{c3}\}_{k=1}^n$, our goal is to infer modality $c$ for dataset $\mathcal{D}^1$. To achieve this, we break down into two steps: 1) first, we leverage $\mathcal{D}^1$ and $\mathcal{D}^2$ to infer the modality $b$, 2) then we use the pseudo embeddings of $b$ in $\mathcal{D}^1$ as a pivot modality shared with $\mathcal{D}^3$ to infer the target modality $c$. Formally, we first capture the cross-modal and cross-dataset relationships:
\begin{align}
    \mathbf{W}^{a\text{-}b}&=\mathbf{F}^{b2}{\mathbf{F}^{a2}}^+,\\
    \mathbf{W}^{2\text{-}1}&=\mathbf{F}^{a1}{\mathbf{F}^{a2}}^+.
\end{align}
Then, we extrapolate modality $b$ in $\mathcal{D}^1$:
\begin{align}
    \tilde{\mathbf{F}}^{b1}_{\text{X-mod}}&=\mathbf{F}^{b2}{\mathbf{F}^{a2}}^+\mathbf{F}^{a1},\\
    \tilde{\mathbf{F}}^{b1}_{\text{X-data}}&=\mathbf{F}^{a1}{\mathbf{F}^{a2}}^+\mathbf{F}^{b2}.
\end{align}
During training, we employ the following Frobenius regularization:
\begin{equation}
\mathcal{L}_{\text{Fro}}(b1)=\|\tilde{\mathbf{F}}^{b1}_{\text{X-mod}}-\tilde{\mathbf{F}}^{b1}_{\text{X-data}}\|_F^2.
\end{equation}
Further, we use $\tilde{\mathbf{F}}^{b1}_{\text{X-data}}$ as the pseudo embeddings of $b$, and then having a embedding set of $\mathcal{D}^1$ as $\{f_i^{a1}, f_i^{b1}\}_{i=1}^l$. Combined with the embedding set of  $\mathcal{D}^3$: $\{f_k^{b3}, f_k^{c3}\}_{k=1}^n$, we have formulate the standard training process of BrokenBind.

Similarly, we first capture cross-modal and cross-dataset relationships, then produce pseudo embeddings:
\begin{align}
    &\mathbf{W}^{b\text{-}c}=\mathbf{F}^{c3}{\mathbf{F}^{b3}}^+,\\
    &\mathbf{W}^{3\text{-}1}=\tilde{\mathbf{F}}^{b1}_{\text{X-data}}{\mathbf{F}^{b3}}^+,\\
    &\tilde{\mathbf{F}}^{c1}_{\text{X-mod}}=\mathbf{F}^{c3}{\mathbf{F}^{b3}}^+\tilde{\mathbf{F}}^{b1}_{\text{X-data}}=\mathbf{F}^{c3}{\mathbf{F}^{b3}}^+\mathbf{F}^{a1}{\mathbf{F}^{a2}}^+\mathbf{F}^{b2},\\
    &\tilde{\mathbf{F}}^{c1}_{\text{X-data}}=\tilde{\mathbf{F}}^{b1}_{\text{X-data}}{\mathbf{F}^{b3}}^+\mathbf{F}^{c3}=\mathbf{F}^{a1}{\mathbf{F}^{a2}}^+\mathbf{F}^{b2}{\mathbf{F}^{b3}}^+\mathbf{F}^{c3}.
\end{align}
The above pseudo embeddings can effectively extrapolate the target modality by \textbf{only leveraging the available modality data}. Hence, we can have the following Frobenius regularization:
\begin{equation}
\mathcal{L}_{\text{Fro}}(c1)=\|\tilde{\mathbf{F}}^{c1}_{\text{X-mod}}-\tilde{\mathbf{F}}^{c1}_{\text{X-data}}\|_F^2.
\end{equation}
After obtaining $\tilde{\mathbf{F}}^{c1}_{\text{X-data}}$, we can apply our MOX objective as follows:
\begin{equation}
\begin{aligned}
    \mathcal{L}_{\text{MOX}}(c1)=&-\frac{1}{B}\sum_{i=1}^B\log\left[\frac{\exp(\langle f^{a1}_i, \tilde{f}^{c1}_i\rangle/\tau)}{\sum_{j=1}^B\exp(\langle f^{a1}_i, \tilde{f}^{c1}_j\rangle/\tau)}\right].
\end{aligned}
\end{equation}
Note that this objective only contains one pair of contrastive learning because $\mathcal{D}^1$ only has one modality. Therefore, our final learning objective can be written as:
\begin{equation}
\mathcal{L}=\mathcal{L}_{\text{CLIP}}+\mathcal{L}_{\text{Cons}}+\mathcal{L}_{\text{Fro}}+\mathcal{L}_{\text{MOX}},
\end{equation}
where $\mathcal{L}_{\text{Fro}}=\mathcal{L}_{\text{Fro}}(b1)+\mathcal{L}_{\text{Fro}}(c1)$. Similarly to the two-dataset scenario, we first conduct pre-training without the MOX objective, then apply the MOX objective for final training.

\section{Additional Experiments}
In this section, we conduct additional experiments to further evaluate the effectiveness of BrokenBind. 

\begin{figure}[h]
    \centering
    \includegraphics[width=0.5\linewidth]{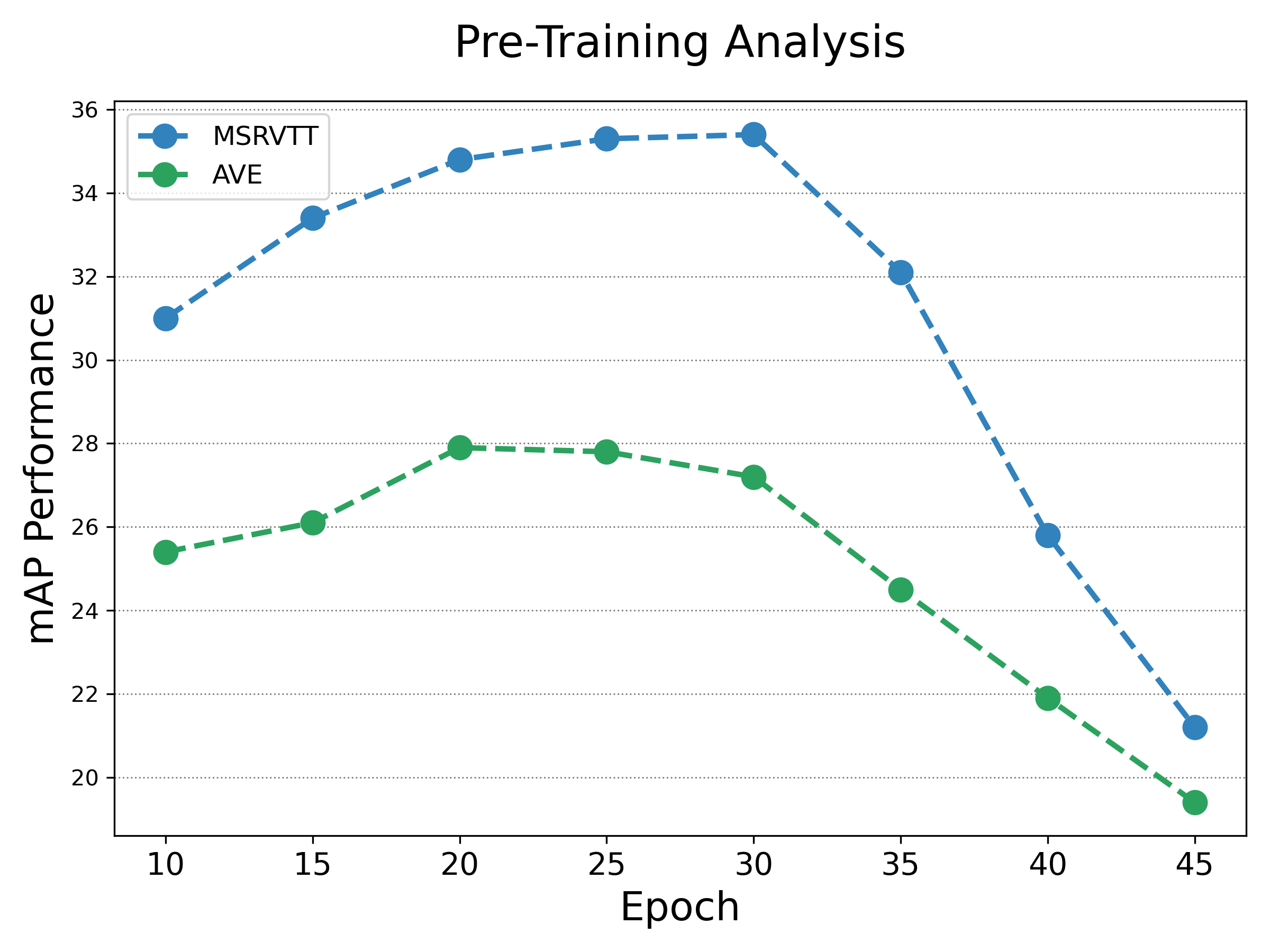}
    \caption{Pre-training analysis by varying the number of pretraining epochs.}
    \label{fig:pre-training}
\end{figure}

First, we demonstrate the performance of consistency pre-training. Specifically, we set 25 training epochs for pre-training in the main paper, here we analyze the performance of BrokenBind when the number of pre-training epochs varies. The results on MSRVTT and AVE datasets are shown in Figure \ref{fig:pre-training}. We can see that both when pre-training for too short and too long would achieve suboptimal performance. When training without consistency pre-training, \textit{i.e.}, using MOX objective the whole time, the performance is limited because the inconsistent embedding space would hinder the effectiveness of modality extrapolation. Moreover, when only using consistency pre-training, \textit{i.e.}, without MOX objective, the performance degenerates to CyCLIP~\cite{goel2022cyclip} which cannot infer unseen modalities, thus leading to poor performance. We can see that when choosing the pre-training epoch around 25 achieves the best performance on both datasets.

\begin{table}[h]
\centering
\renewcommand{\arraystretch}{1.0} 
\setlength{\tabcolsep}{3pt} 
\small 
\caption{Performance of binding emergent modalities across datasets. Model using VLIP combined with CLAP, and modalities include vi, te, and au.}
\begin{tabular}{@{}lcccc@{}}
\toprule[1.2pt] 
\multirow{2}{*}{Method} & \multicolumn{2}{c}{AVE $\rightarrow$ MSRVTT} & \multicolumn{2}{c}{MSRVTT $\rightarrow$ AVE} \\
\cmidrule(lr){2-3} \cmidrule(lr){4-5}
& \footnotesize au$\rightarrow$te$\rightarrow$vi & \footnotesize vi$\rightarrow$te$\rightarrow$au & 
\footnotesize au$\rightarrow$te$\rightarrow$vi & \footnotesize vi$\rightarrow$te$\rightarrow$au \\
\midrule
VCLIP+CLAP              & 5.1 & 6.2 & 4.2 & 3.8 \\
VCLIP+CLAP+FT           & 8.4 & 9.6 & 7.0 & 6.2 \\
C-MCR                   & 8.5 & 9.8 & 6.7 & 5.6 \\
Ex-MCR                  & 9.0 & 10.2 & 7.1 & 5.8 \\
VCLIP+CLAP+BB           & \textbf{17.6} & \textbf{17.4} & \textbf{16.8} & \textbf{18.1} \\
\bottomrule[1.2pt]
\end{tabular}
\label{tab:comparison_app}
\end{table}

Further, we consider two additional baselines C-MCR~\cite{wang2023connecting} and Ex-MCR~\cite{wang2023extending} which both aim to combine two different modality bindings to achieve cross-modal generalization. The result is shown in Table \ref{tab:comparison_app}. We can see that our method can still significantly surpass the two baselines because the methodology still directly aligns two modalities together without considering the bias between datasets. Therefore, the effectiveness of BrokenBind in tackling modality binding beyond dataset boundaries is again well-justified.

\begin{figure}[h]
    \centering
    \includegraphics[width=\linewidth]{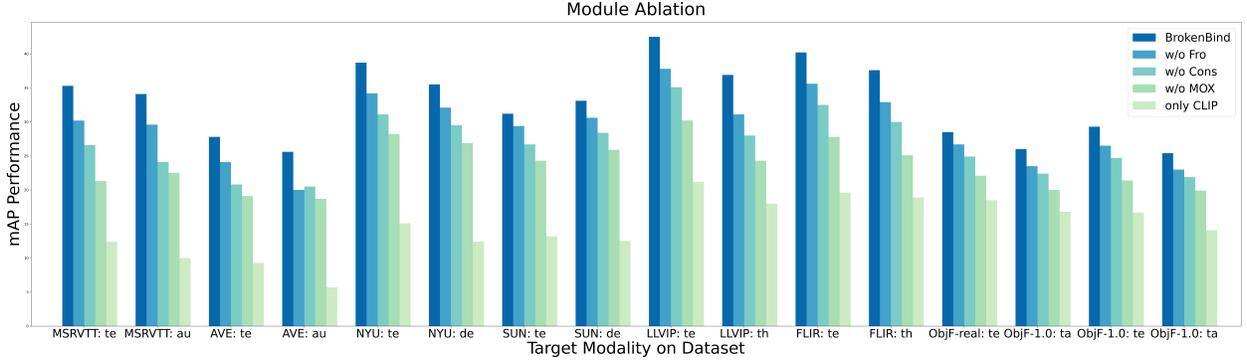}
    \caption{Ablation study on different module settings.}
    \label{fig:app_ablation}
\end{figure}

Moreover, we conduct an additional ablation study on each module of BrokenBind under various scenarios, as shown in Figure \ref{fig:app_ablation}. We can see that the importance of MOX is clear under all circumstances, which again justifies the effectiveness of the extrapolation process.

\begin{figure}[h]
    \centering
    \includegraphics[width=\linewidth]{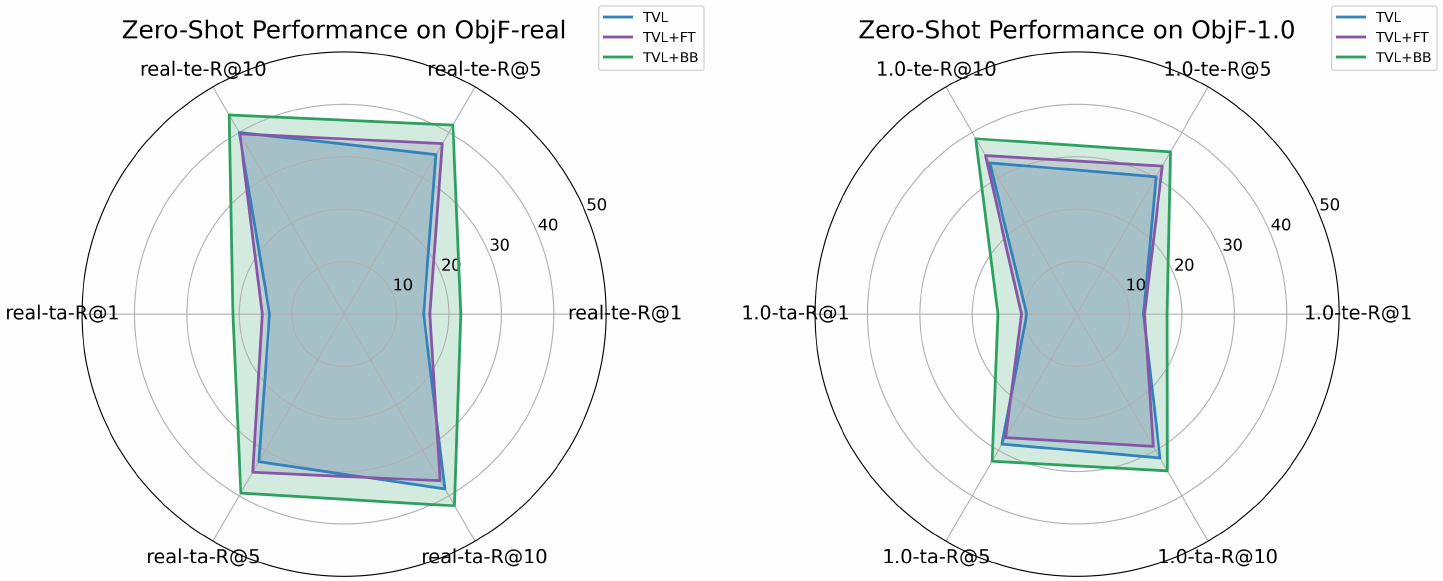}
    \caption{Zero-shot learning on ObjF-real and ObjF-1.0 datasets with different target modalities.}
    \label{fig:zero_shot}
\end{figure}

Finally, we study the zero-shot learning performance of BrokenBind compared to the TVL backbone~\cite{fu2024touch}. Specifically, we pre-train our model using the ObjF-2.0 dataset and VGG-S dataset, which contain ta, te, vi, au, and te, vi, au modalities, respectively. We consider target modalities as te and ta, and use vi modality as the  pivot modality. To testify the zero-shot performance, we use ObjF-real and ObjF-1.0 datasets and show the performance in Figure \ref{fig:zero_shot}. We can see that compared to the TVL backbone, our BrokenBind can further enhance the zero-shot performance in all situations, which demonstrates that our BrokenBind can be effectively applied to large-scale fine-tuning and combine separate datasets to achieve modality binding.

\end{document}